\documentclass[11pt,letterpaper]{article}

\usepackage{naaclhlt2016}

\usepackage{amsmath,amsfonts}
\usepackage{amsthm}
\usepackage{times}
\usepackage{xcolor}
\usepackage{url}
\usepackage{balance}
\usepackage{booktabs}

\usepackage{booktabs}

\usepackage{array}
\newcolumntype{x}[1]{>{\centering\arraybackslash\hspace{0pt}}p{#1}}

\naaclfinalcopy

\usepackage{cleveref}
\Crefformat{section}{\S#2#1#3}

\usepackage{tikz}
\usetikzlibrary{automata,positioning}

\usepackage{subcaption}
\usepackage{rotating}
\usepackage{xcolor}
\usepackage{enumitem}
\usepackage{tikz-qtree}

\definecolor{myblue}{rgb}{0.25,0.41,0.88}
\definecolor{mygreen}{rgb}{0.15,0.55,0.11}
\definecolor{myred}{rgb}{0.88,0.41,0.25}

\usepackage{algorithm,algpseudocode}
\usepackage{xspace}
\usepackage{amsmath}

\usepackage{xargs} 
\usepackage[colorinlistoftodos,prependcaption,textsize=tiny]{todonotes}
\usepackage{marginnote}
\newcommand \ignore[1]{}
\newcommandx{\todor}[1]{\textcolor{red}{SR: #1}}
\newcommandx{\todoi}[1]{\textcolor{red}{#1}}
\newcommandx{\todon}[1]{\textcolor{cyan}{SN: #1}}
\newcommandx{\todoc}[1]{\textcolor{blue}{SC: #1}}

\newcommand \gp{\textsc{GraphParser}\xspace}

\newcommand \webq{WebQuestions\xspace}

\newcommand{\argmax}{\operatornamewithlimits{arg\,max}}

\newcommand{\lconp}{G_{\mathrm{syn}}^{\prime}}
\newcommand{\lcon}{G_{\mathrm{syn}}}
\newcommand{\lhyb}{G_{\mathrm{layered}}}
\newcommand{\lpar}{G_{\mathrm{par}}}

\newfont{\msym}{msbm10}
\newcommand{\reals}{\mbox{\msym R}}

\newcommand{\xione}{t^{(i)}}

\newcommand{\aione}{a^{(i)}}

\newcommand{\oi}{o^{(i)}}

\newcommand{\p}{{\cal P}}
\newcommand{\internal}{{\cal I}}
\newcommand{\n}{{\cal N}}

\title{Paraphrase Generation from Latent-Variable PCFGs for Semantic Parsing}

\author{Shashi Narayan, Siva Reddy and Shay B. Cohen \\ School of
  Informatics, University of Edinburgh \\ 10 Crichton Street,
  Edinburgh, EH8 9LE, UK \\ \small{\texttt{shashi.narayan@ed.ac.uk},
    \texttt{siva.reddy@ed.ac.uk}, \texttt{scohen@inf.ed.ac.uk} }}

\begin{document}

\maketitle

\begin{abstract}
  One of the limitations of semantic parsing approaches to open-domain
  question answering is the lexicosyntactic gap between natural
  language questions and knowledge base entries -- there are many ways
  to ask a question, all with the same answer.  In this paper we
  propose to bridge this gap by generating paraphrases of the input
  question with the goal that at least one of them will be correctly
  mapped to a knowledge-base query. We introduce a novel
  grammar model for paraphrase generation that does not require any
  sentence-aligned paraphrase corpus. Our key idea is to leverage the
  flexibility and scalability of latent-variable probabilistic
  context-free grammars to sample paraphrases. We do an extrinsic
  evaluation of our paraphrases by plugging them into a semantic
  parser for Freebase. Our evaluation experiments on the WebQuestions
  benchmark dataset show that the performance of the semantic parser
  improves over strong baselines.
\end{abstract}

\section{Introduction}
\label{sec:intro}

Semantic parsers map sentences onto logical forms that can be used to
query databases \cite{zettlemoyer_learning_2005,wong_learning_2006},
instruct robots \cite{chen_learning_2011}, extract information
\cite{krishnamurthy_weakly_2012}, or describe visual scenes
\cite{matuszek_joint_2012}. In this paper we consider the problem of
semantically parsing questions into Freebase logical forms for the
goal of question answering.  Current systems accomplish this by
learning task-specific grammars \cite{berant_semantic_2013},
strongly-typed CCG grammars
\cite{kwiatkowski_scaling_2013,reddy_largescale_2014}, or neural
networks without requiring any grammar \cite{yih_semantic_2015}. These
methods are sensitive to the words used in a question and their word
order, making them vulnerable to unseen words and
phrases. Furthermore, mismatch between natural language and Freebase
makes the problem even harder. For example, Freebase expresses the
fact that \textit{``Czech is the official language of Czech Republic''}
(encoded as a graph), whereas to answer a question like \textit{``What
  do people in Czech Republic speak?''} one should infer
\textit{people in Czech Republic} refers to \textit{Czech Republic}
and \textit{What} refers to the \textit{language} and \textit{speak}
refers to the predicate \textit{official language}.

We address the above problems by using paraphrases of the original
question. Paraphrasing has shown to be promising for semantic parsing
\cite{fader_paraphrasedriven_2013,berant_semantic_2014,wang_building_2015}.
We propose a novel framework for paraphrasing using latent-variable
PCFGs (L-PCFGs). Earlier approaches to paraphrasing used phrase-based
machine translation for text-based QA \cite{Duboue2006,Riezler2007},
or hand annotated grammars for KB-based QA
\cite{berant_semantic_2014}. We find that phrase-based statistical
machine translation (MT) approaches mainly produce lexical
paraphrases without much syntactic diversity, whereas our
grammar-based approach is capable of producing both lexically and
syntactically diverse paraphrases. Unlike MT based approaches, our
system does not require aligned parallel paraphrase corpora. In
addition we do not require hand annotated grammars for paraphrase
generation but instead learn the grammar directly from a large scale
question corpus.

The main contributions of this paper are two fold. First, we present
an algorithm (\S\ref{sec:paragen}) to generate paraphrases using
latent-variable PCFGs. We use the spectral method of
\newcite{narayan-15} to estimate L-PCFGs on a large scale question
treebank. Our grammar model leads to a robust and an efficient system
for paraphrase generation in open-domain question answering. While
CFGs have been explored for paraphrasing using bilingual parallel
corpus \cite{ppdb}, ours is the first implementation of CFG that uses
only monolingual data. Second, we show that generated paraphrases can
be used to improve semantic parsing of questions into Freebase logical
forms (\S\ref{sec:qaframework}). We build on a strong baseline of
\newcite{reddy_largescale_2014} and show that our grammar model
competes with MT baseline even without using any parallel paraphrase
resources.


\vspace{-0.1cm}
\section{Paraphrase Generation Using Grammars}
\label{sec:paragen}
\vspace{-0.1cm}

Our paraphrase generation algorithm is based on a model in the form of
an L-PCFG.  L-PCFGs are PCFGs where the nonterminals are refined with
latent states that provide some contextual information about each node
in a given derivation.  L-PCFGs have been used in various ways, most
commonly for syntactic parsing
\cite{prescher-05,matsuzaki-2005,petrov-2006,cohen-13,narayan-15,narayan-16b}.

In our estimation of L-PCFGs, we use the spectral method of
\newcite{narayan-15}, instead of using EM, as has been used in the
past by \newcite{matsuzaki-2005} and \newcite{petrov-2006}. The
spectral method we use enables the choice of a set of feature
functions that indicate the latent states, which proves to be useful
in our case. It also leads to sparse grammar estimates and compact
models.

The spectral method works by identifying feature functions for
``inside'' and ``outside'' trees, and then clusters them into latent
states.  Then it follows with a maximum likelihood estimation step,
that assumes the latent states are represented by clusters obtained
through the feature function clustering.  For more details about these
constructions, we refer the reader to \newcite{cohen-13} and
\newcite{narayan-15}.

The rest of this section describes our paraphrase generation
algorithm.

\vspace{-0.1cm}
\subsection{Paraphrases Generation Algorithm}
\label{subsec:samplepara}

\ignore{
\begin{figure}
  \begin{footnotesize}
    \begin{algorithmic}[1]

      \State \textbf{Input}: An input question $q$, an L-PCFG
      $\lcon$ with parameters $(\n, \internal, \p, m_{con}, n,
      \pi_{con}, t_{con}, q_{con})$, a paraphrase classifier
      $C:(s_1,s_2)\rightarrow[0,1]$ and an integer $M$. $\lcon$ is
      the grammar of \newcite{narayan-15} for constituency
      parsing. $C$ checks if two sentences $s_1$ and $s_2$ are
      paraphrases. $M$ is the number of sampling to be
      done. Optionally, the algorithm may take the Paraphrase Database
      $D_{ppdb}$ or an hybrid L-PCFG $\lhyb$.

      \State \textbf{Output}: A set of paraphrases $P_q$ to the input
      question $q$.

      \State \textbf{Algorithm}:

      \State initialize $P_s \leftarrow \emptyset$

      \State build a word lattice $W_q$ for $q$, optionally using
      $D_{ppdb}$ or $\lhyb$

      \State extract a smaller L-PCFG $\lconp$ from $\lconp$ for
      $W_q$

      \For{i = 1}{ $M$}
      \State sample a sentence $q'$ from $\lconp$
      \If{$C(q,q')==1$}
      \State $P_q = P_q \cup {q'}$ 
      \EndIf
      \EndFor

    \end{algorithmic}
  \end{footnotesize}
  \caption{The paraphrase generation algorithm.}
  \label{fig:paragenalgo}
\end{figure}
}

\begin{figure*}[htbp]
  \centering
  \includegraphics[width=\textwidth]{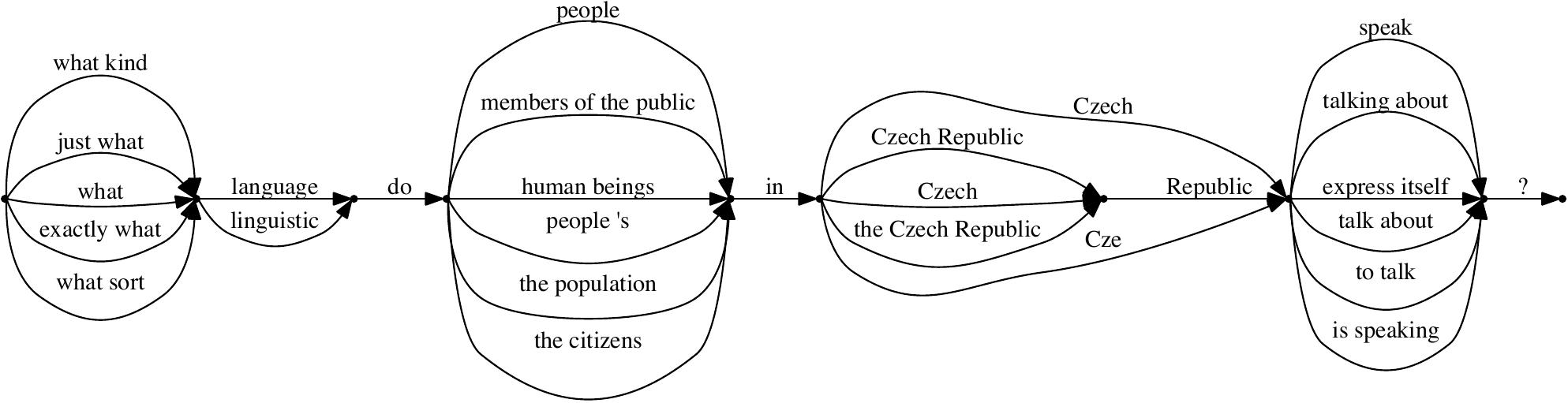}
  \caption{\small An example word lattice for the question {\it What language
      do people in Czech Republic speak?} using the lexical and
    phrasal rules from the PPDB. 
  }
  \label{fig:wordlattice}
  \vspace{-0.4cm}
\end{figure*}


We define our paraphrase generation task as a sampling problem from an
L-PCFG $\lcon$, which is estimated from a large corpus of parsed
questions. Once this grammar is estimated, our algorithm follows a
pipeline with two major steps.

We first build a word lattice $W_q$ for the input question
$q$.\footnote{Word lattices, formally weighted finite state automata,
  have been used in previous works for paraphrase generation
  \cite{langkilde1998,barzilay2003,pang2003,quirk-2004}. We use an
  unweighted variant of word lattices in our algorithm.}
We use the lattice to constrain our paraphrases to a specific choice
of words and phrases that can be used.
Once this lattice is created, a grammar $\lconp$ is then extracted
from $\lcon$.  This grammar is constrained to the lattice.

\ignore{
In this setting, we use the PPDB \cite{ppdb} to build word
lattices. For an input question $q$, we extract high-precision rules
from the PPDB\footnote{For our experiments, we extract rules from the
  PPDB-Small to maintain the high precision \cite{ppdb}.} which match
the pattern in $q$. We use these rules to build the word lattice. We
use both lexical and phrasal paraphrase rules. We refer this system by
\textbf{PPDB} in future reference. Our approach is similar to
\newcite{langkilde1998} where they use WordNet-based hand-crafted
rules to build word lattices. Figure~\ref{fig:wordlattice} shows an
example word lattice for the question {\it What language do people in
  Czech Republic speak?} using the rules from the PPDB.
}

We experiment with three ways of constructing word lattices: na\"{i}ve
word lattices representing the words from the input question only,
word lattices constructed with the Paraphrase Database \cite{ppdb} and
word lattices constructed with a bi-layered L-PCFG, described in
\S\ref{subsec:hybrid}. For example, Figure~\ref{fig:wordlattice} shows
an example word lattice for the question {\it What language do people
  in Czech Republic speak?} using the lexical and phrasal rules from
the PPDB.\footnote{For our experiments, we extract rules from the
  PPDB-Small to maintain the high precision \cite{ppdb}.}


\ignore{
We then extract a smaller grammar $\lconp$ from $\lcon$ to
sample new questions. $\lconp$ is the compact
representation of $\lcon$ for $W_q$. We describe this pruning step
in \S\ref{subsec:pruning}. 
}

Once $\lconp$ is generated, we sample paraphrases of the input
question $q$. These paraphrases are further filtered with a classifier
to improve the precision of the generated paraphrases.

\ignore{The sampling algorithm, described in \S\ref{subsec:sampling},
  is not guaranteed to generate paraphrases. We use a classifier,
  described in \S\ref{subsec:classifier}, to select paraphrases among
  the sampled questions (step 9).  }


\paragraph{L-PCFG Estimation}
We train the L-PCFG $\lcon$ on the Paralex corpus
\cite{fader_paraphrasedriven_2013}.
Paralex is a large monolingual parallel corpus, containing 18 million
pairs of question paraphrases with 2.4M distinct questions in the
corpus.  It is suitable for our task of generating paraphrases since
its large scale makes our model robust for open-domain questions. We
construct a treebank by parsing 2.4M distinct questions from Paralex
using the BLLIP parser \cite{charniak-johnson:2005:ACL}.\footnote{We
  ignore the Paralex alignments for training $\lcon$.}

Given the treebank, we use the spectral algorithm of
\newcite{narayan-15} to learn an L-PCFG for constituency parsing to
learn $\lcon$. We follow \newcite{narayan-15} and use the same feature
functions for the inside and outside trees as they use, capturing contextual
syntactic information about nonterminals. We refer the reader to
\newcite{narayan-15} for more detailed description of these features.
In our experiments, we set the number of latent states to 24.

Once we estimate $\lcon$ from the Paralex corpus, we restrict it for
each question to a grammar $\lconp$ by keeping only the rules that
could lead to a derivation over the lattice. This step is similar to
lexical pruning in standard grammar-based generation process to avoid
an intermediate derivation which can never lead to a successful
derivation \cite{koller-2002,nargar-2012}.

\paragraph{Paraphrase Sampling} 

Sampling a question from the grammar $\lconp$ is done by recursively
sampling nodes in the derivation tree, together with their latent
states, in a top-down breadth-first fashion. Sampling from the pruned
grammar $\lconp$ raises an issue of oversampling words that are more
frequent in the training data. To lessen this problem, 
we follow a {\it controlled sampling} approach where sampling is
guided by the word lattice $W_q$. Once a word $w$ from a path $e$ in
$W_q$ is sampled, all other parallel or conflicting paths to $e$ are
removed from $W_q$. For example, generating for the word lattice in
Figure \ref{fig:wordlattice}, when we sample the word {\it citizens},
we drop out the paths {\it ``human beings''}, {\it ``people's''}, {\it
  ``the population''}, {\it ``people''} and {\it ``members of the
  public''} from $W_q$ and accordingly update the grammar. The
controlled sampling ensures that each sampled question uses words from
a single start-to-end path in $W_q$. For example, we could sample a
question {\it what is Czech Republic 's language?} by sampling words
from the path {\it (what, language, do, people 's, in, Czech,
  Republic, is speaking, ?)} in Figure
\ref{fig:wordlattice}. \ignore{We note that the controlled sampling
  may lead to a failed derivation. The derivation of a tree could
  introduce an interminal which will fail to complete because some
  words from in $W_q$ were pruned out since the interminal was first
  introduced. Breadth-first search makes it faster to identify an
  unsuccessful derivation.} We repeat this sampling process to
generate multiple potential paraphrases.

\ignore{
Our generation algorithm differs from existing literature
\cite{langkilde1998,barzilay2003,pang2003,quirk-2004} on the usage of
word lattices to generate paraphrases. Existing systems often use
alternative paths between two nodes in word lattices to generate
paraphrases. In contrast, we sample words from a path in the word
lattice using a grammar regardless of their order in the path. For
example, we could sample a question {\it what is Czech Republic 's
  language?} by sampling words from the path {\tt (what, language, do,
  people 's, in, Czech, Republic, is speaking, ?)} in Figure
\ref{fig:wordlattice}.
}

The resulting generation algorithm has multiple advantages over
existing grammar generation methods. First, the sampling from an
L-PCFG grammar lessens the lexical ambiguity problem evident in
lexicalized grammars such as tree adjoining grammars \cite{nargar-2012} and combinatory
categorial grammars \cite{white-2004}. 
Our grammar is not lexicalized, only unary context-free rules are
lexicalized. Second, the top-down sampling restricts the combinatorics
inherent to bottom-up search \cite{Shieber1990}. Third, we do not
restrict the generation by the order information in the input. The
lack of order information in the input often raises the high
combinatorics in lexicalist approaches \cite{kay-1996}. In our case,
however, we use sampling to reduce this problem, and it allows us to
produce syntactically diverse questions. And fourth, we impose no
constraints on the grammar thereby making it easier to maintain
bi-directional (recursive) grammars that can be used both for parsing
and for generation \cite{shieber1988}.



\begin{figure*}[htbp]
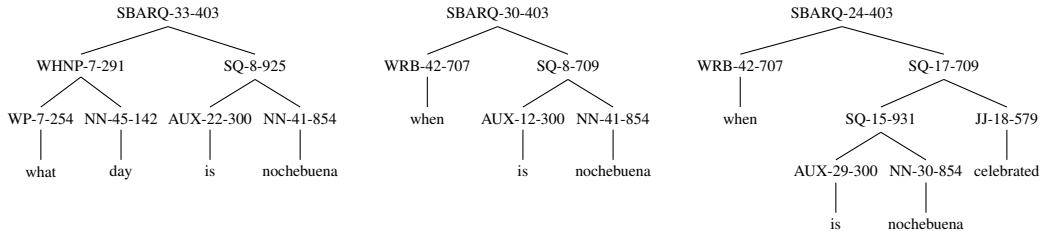

  \begin{center}
    \begin{tiny}
    \hspace{-0.7cm}  \begin{tabular}{lll}
      \tikzset{level distance=20pt, sibling distance=0pt}
        \Tree [.SBARQ-33-403 [.WHNP-7-291 [.WP-7-254 what ] [.NN-45-142 day ] ] [.SQ-8-925 [.AUX-22-300 is ] [.NN-41-854 nochebuena ] ] ]
        &
        \tikzset{level distance=20pt, sibling distance=0pt}
        \Tree [.SBARQ-30-403 [.WRB-42-707 when ] [.SQ-8-709 [.AUX-12-300 is ] [.NN-41-854 nochebuena ] ] ]
        &
        \tikzset{level distance=20pt, sibling distance=0pt}
        \Tree [.SBARQ-24-403 [.WRB-42-707 when ] [.SQ-17-709 [.SQ-15-931 [.AUX-29-300 is ] [.NN-30-854 nochebuena ] ] [.JJ-18-579 celebrated ] ] ]
      \end{tabular}
    \end{tiny}
    \end{center}
    \caption{\small Trees used for bi-layered L-PCFG training. The questions
      {\it what day is nochebuena}, {\it when is nochebuena} and {\it
        when is nochebuena celebrated} are paraphrases from the
      Paralex corpus. Each nonterminal is decorated with a syntactic
      label and two identifiers, e.g., for WP-7-254, WP is the
      syntactic label assigned by the BLLIP parser, 7 is the syntactic
      latent state, and 254 is the semantic latent state.}
  \label{fig:hybridtrees}
  \vspace{-0.3cm}
\end{figure*}


\subsection{Bi-Layered L-PCFGs}
\label{subsec:hybrid}

As mentioned earlier, one of our lattice types is based on bi-layered
PCFGs introduced here.

In their traditional use, the latent states in L-PCFGs aim to capture
syntactic information. We introduce here the use of an L-PCFG with two
layers of latent states: one layer is intended to capture the usual
syntactic information, and the other aims to capture semantic and
topical information by using a large set of states with specific
feature functions.\footnote{For other cases of separating syntax from
  semantics in a similar way, see \newcite{mitchell-15}.}

\ignore{
We define an L-PCFG $\lhyb$ with parameters $(\n, \internal, \p,
m_{hyb}, n, \pi_{hyb}, t_{hyb}, q_{hyb})$ as a hybrid between an
L-PCFG $\lcon$ for constituency parsing with parameters $(\n,
\internal, \p, m_{con}, n, \pi_{con}, t_{con}, q_{con})$ and an L-PCFG
$\lpar$ for paraphrasing with parameters $(\n, \internal, \p,
m_{par}, n, \pi_{par}, t_{par}, q_{par})$. We have shown how we train
$\lcon$ in \S\ref{subsec:samplegrammar}. Here, we describe how we
train $\lpar$ on the Paralex corpus, and subsequently $\lhyb$.

$\lpar$ aims to capture the paraphrase information about the
nonterminals in the grammar. The intuition behind this grammar is that
the nonterminals, situated in the context of paraphrases and yielding
paraphrases, should observe same latent state. For example, in Figure
\ref{fig:hybridtrees}, the nonterminal {\tt SBARQ} in all three trees
yields paraphrases {\it what day is nochebuena}, {\it when is
  nochebuena} and {\it when is nochebuena celebrated}, hence {\tt
  SBARQ} in all three trees should observe same latent state. Given a
parallel corpus of aligned paraphrase pairs where these three
sentences are aligned to each other, {\tt SBARQ}s could learn this
paraphrase information from their word alignments. With this
intuition, we define the inside feature functions $\phi$ as a mapping
from an inside tree to a bag of words consisting of the direct and
aligned yields of the inside tree. Similarly, we define the outside
feature function $\psi$ for outside trees.

}

To create the bi-layered L-PCFG, we again use the spectral algorithm
of \newcite{narayan-15} to estimate a grammar $\lpar$ from the Paralex
corpus. We use the word alignment of paraphrase question pairs in
Paralex to map inside and outside trees of each nonterminals in the
treebank to bag of word features. The number of latent states we use
is 1,000.

Once the two feature functions (syntactic in $\lcon$ and semantic in
$\lpar$) are created, each nonterminal in the training treebank is
assigned two latent states (cluster identifiers). Figure
\ref{fig:hybridtrees} shows an example annotation of trees for three
paraphrase questions from the Paralex corpus. We compute the
parameters of the bi-layered L-PCFG $\lhyb$ with a simple frequency
count maximum likelihood estimate over this annotated treebank. As
such, $\lhyb$ is a combination of $\lcon$ and $\lpar$, resulting in
24,000~latent states (24~syntactic x 1000~semantic).



Consider an example where we want to generate paraphrases for the
question {\it what day is nochebuena}. Parsing it with $\lhyb$ will
lead to the leftmost hybrid structure as shown in Figure
\ref{fig:hybridtrees}. The assignment of the first latent states for
each nonterminals ensures that we retrieve the correct syntactic
representation of the sentence. Here, however, we are more interested
in the second latent states assigned to each nonterminals which
capture the paraphrase information of the sentence at various levels.
For example, we have a unary lexical rule {\tt (NN-*-142 day)}
indicating that we observe {\it day} with {\tt NN} of the paraphrase
type {\tt 142}. We could use this information to extract unary rules
of the form {\tt (NN-*-142 $w$)} in the treebank that will generate
words $w$ which are paraphrases to {\it day}. Similarly, any node {\tt
  WHNP-*-291} in the treebank will generate paraphrases for {\it what
  day}, {\tt SBARQ-*-403}, for {\it what day is nochebuena}. This way
we will be able to generate paraphrases {\it when is nochebuena} and
{\it when is nochebuena celebrated} as they both have {\tt
  SBARQ-*-403} as their roots.\footnote{We found out that our $\lpar$
  grammar is not fine-grained enough and often merges different
  paraphrase information into the same latent state. This problem is
  often severe for nonterminals at the top level of the bilayered
  tree. \ignore{For example, consider a tree {\tt (SBARQ-1-403
      (WRB-23-103 where) (SQ-22-809 (SQ-21-910 (AUX-10-866 was)
      (NP-24-60 (NNP-21-567 gabuella) (NNP-21-290 montez)))
      (VBN-29-682 born)))} for the sentence {\it where was gabuella
      montez born} in the treebank. The root of the tree is assigned
    {\tt SBARQ-*-403} but it is not a paraphrase to the sentence {\it
      what day is nochebuena}. } Hence, we rely only on unary lexical
  rules (the rules that produce terminal nodes) to extract paraphrase
  patterns in our experiments.}


To generate a word lattice $W_q$ for a given question $q$, we parse
$q$ with the bi-layered grammar $\lhyb$. For each rule of the form
$X$-$m_1$-$m_2 \rightarrow w$ in the bi-layered tree with $X \in \p$,
$m_1 \in \{ 1, \ldots, 24 \}$, $m_2 \in \{ 1, \ldots, 1000 \}$ and $w$
a word in $q$, we extract rules of the form $X$-$*$-$m_2 \rightarrow
w'$ from $\lhyb$ such that $w' \neq w$. For each such $(w, w')$, we
add a path $w'$ parallel to $w$ in the word lattice.

\subsection{Paraphrase Classification}
\label{subsec:classifier}

Our sampling algorithm overgenerates paraphrases which are
incorrect. To improve its precision, we build a binary classifier to
filter the generated paraphrases. We randomly select 100 distinct
questions from the Paralex corpus and generate paraphrases using our
generation algorithm with various lattice settings. We randomly select
1,000 pairs of input-sampled sentences and manually annotate them as
``correct'' or ``incorrect'' paraphrases.\footnote{We have 154 positive and 846 negative paraphrase pairs.} We train our classifier on
this manually created training data.\footnote{We do not use the
  paraphrase pairs from the Paralex corpus to train our classifier, as
  they do not represent the distribution of our sampled paraphrases
  and the classifier trained on them performs poorly.}
We follow \newcite{madnani2012}, who used MT metrics for paraphrase
identification, and experiment with 8 MT metrics as features for our
binary classifier.
In addition, we experiment with a binary feature which checks if the
sampled paraphrase preserves named entities from the input sentence.
We use WEKA \cite{hall2009weka} to replicate the classifier of
\newcite{madnani2012} with our new feature.
We tune the feature set for our classifier on the development data.

\section{Semantic Parsing using Paraphrasing}
\label{sec:qaframework}




\begin{figure*}
  \centering
  \begin{minipage}{0.6\textwidth}
    \begin{subfigure}{\textwidth}
      \includegraphics[width=\textwidth]{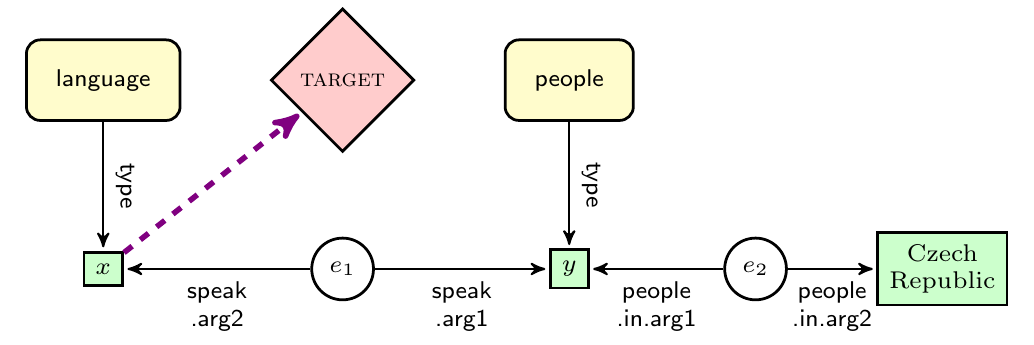}
      \caption{Input sentence: What language do people in Czech Republic speak?}
      \label{fig:originalGraph}
    \end{subfigure} \\
    \begin{subfigure}{0.48\textwidth}
      \addtocounter{subfigure}{+1}
      \vspace{0.4cm}
      \includegraphics[width=\textwidth]{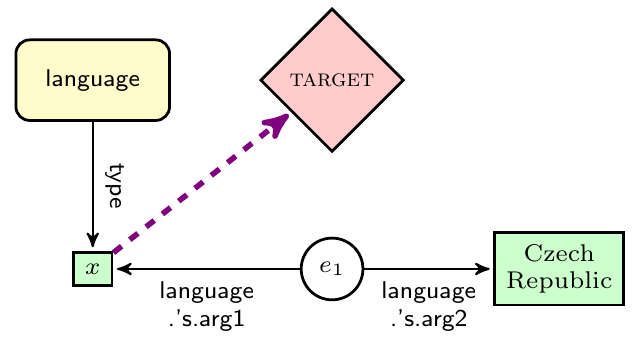}
      \caption{Paraphrase: What is Czech Republic's language?}
      \label{fig:paraphraseGraph2}
    \end{subfigure}
    \begin{subfigure}{0.48\textwidth}
      \includegraphics[width=1.2\textwidth]{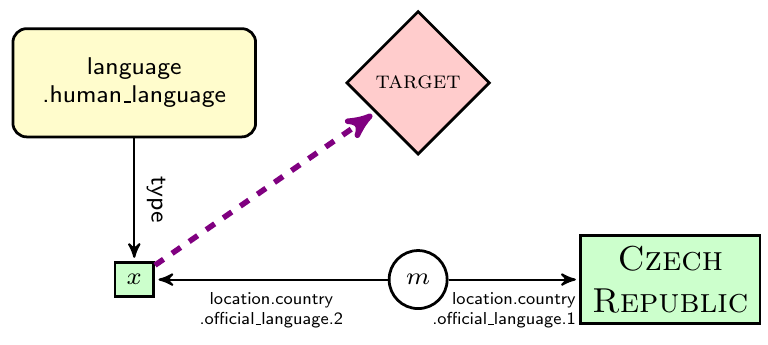}
      \caption{Freebase grounded graph}
      \label{fig:groundedGraph}
    \end{subfigure}
  \end{minipage}
  \hspace{1cm}
  \begin{minipage}{0.32\textwidth}
    \vspace{0.3cm}
    \begin{subfigure}{\textwidth}
      \addtocounter{subfigure}{-3}
      \includegraphics[width=\textwidth]{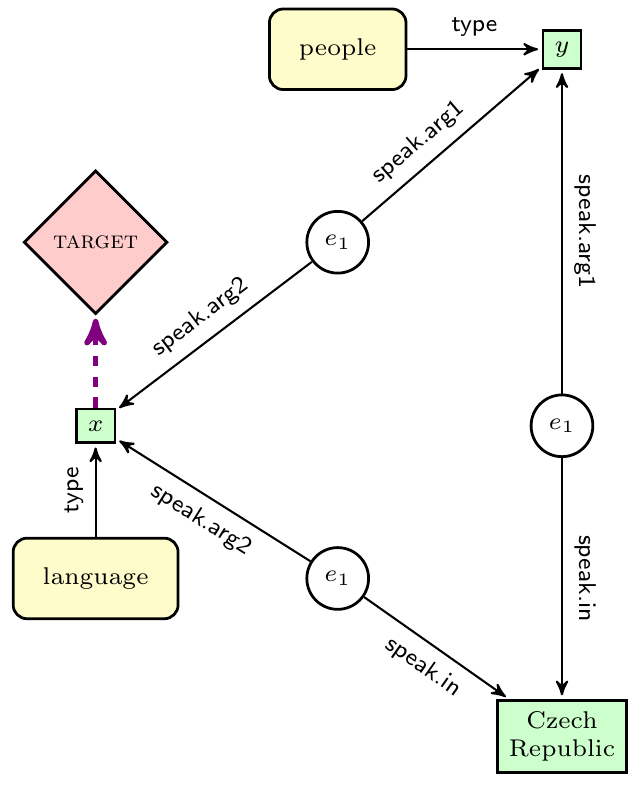}
      \caption{Paraphrase: What language do people speak in Czech Republic?}
      \label{fig:paraphraseGraph1}
    \end{subfigure}
  \end{minipage}
  \caption{\small Ungrounded graphs for an input question and its
    paraphrases along with its correct grounded graph. The green
    squares indicate NL or Freebase entities, the yellow rectangles
    indicate unary NL predicates or Freebase types, the circles
    indicate NL or Freebase events, the edge labels indicate binary NL
    predicates or Freebase relations, and the red diamonds attach to
    the entity of interest (the answer to the question).}
  \label{fig:paraphraseQA}
  \vspace{-0.3cm}
\end{figure*}

In this section we describe how the paraphrase algorithm is used for
converting natural language to Freebase queries. Following
\newcite{reddy_largescale_2014}, we formalize the semantic parsing
problem as a graph matching problem, i.e., finding the Freebase
subgraph (grounded graph) that is isomorphic to the input question
semantic structure (ungrounded graph).

This formulation has a major limitation that can be alleviated by
using our paraphrase generation algorithm. Consider the question
\emph{What language do people in Czech Republic speak?}. The
ungrounded graph corresponding to this question is shown in
Figure~\ref{fig:originalGraph}. The Freebase grounded graph which
results in correct answer is shown in
Figure~\ref{fig:groundedGraph}. Note that these two graphs are
non-isomorphic making it impossible to derive the correct grounding
from the ungrounded graph. In fact, at least 15\% of the examples in
our development set fail to satisfy isomorphic assumption. In order to
address this problem, we use paraphrases of the input question to
generate additional ungrounded graphs, with the aim that one of those
paraphrases will have a structure isomorphic to the correct
grounding. Figure~\ref{fig:paraphraseGraph1} and
Figure~\ref{fig:paraphraseGraph2} are two such paraphrases which can
be converted to Figure~\ref{fig:groundedGraph} as described in
\Cref{sec:groundedGraphs}.

For a given input question, first we build ungrounded graphs from its
paraphrases.  We convert these graphs to Freebase graphs. To learn
this mapping, we rely on manually assembled question-answer pairs. For
each training question, we first find the set of \emph{oracle}
grounded graphs---Freebase subgraphs which when executed yield the
correct answer---derivable from the question's ungrounded
graphs. These oracle graphs are then used to train a structured
perceptron model. These steps are discussed in detail below.

\subsection{Ungrounded Graphs from Paraphrases}
\label{sec:ungroundedGraphs}
We use \gp \cite{reddy_largescale_2014} to convert paraphrases to
ungrounded graphs. This conversion involves three steps: 1) parsing
the paraphrase using a CCG parser to extract syntactic derivations
\cite{lewis_ccg_2014}, 2) extracting logical forms from the CCG
derivations \cite{bos_widecoverage_2004}, and 3) converting the
logical forms to an ungrounded graph.\footnote{Please see
  \newcite{reddy_largescale_2014} for more details.} The ungrounded
graph for the example question and its paraphrases are shown in
Figure~\ref{fig:originalGraph}, Figure~\ref{fig:paraphraseGraph1} and
Figure~\ref{fig:paraphraseGraph2}, respectively.

\subsection{Grounded Graphs from Ungrounded Graphs}
\label{sec:groundedGraphs}
The ungrounded graphs are grounded to Freebase subgraphs by mapping
entity nodes, entity-entity edges and entity type nodes in the
ungrounded graph to Freebase entities, relations and types,
respectively. For example, the graph in Figure~\ref{fig:paraphraseGraph1}
can be converted to a Freebase graph in Figure~\ref{fig:groundedGraph} by
replacing the entity node \textit{Czech Republic} with the Freebase
entity \textsc{CzechRepublic}, the edge
\textit{(speak.arg$_2$,~speak.in)} between $x$ and \textit{Czech
  Republic} with the Freebase relation
\textit{(location.country.official\_language.2,~location.country.official\_language.1)},
the type node \textit{language} with the Freebase type
\textit{language.human\_language}, and the \textsc{target} node
remains intact. The rest of the nodes, edges and types are grounded to
\textit{null}. In a similar fashion, Figure~\ref{fig:paraphraseGraph2} can
be grounded to Figure~\ref{fig:groundedGraph}, but not
Figure~\ref{fig:originalGraph} to Figure~\ref{fig:groundedGraph}. If no paraphrase is isomorphic to the target grounded grounded graph, our grounding fails.

\subsection{Learning}
\label{sec:learning}

We use a linear model to map ungrounded graphs to grounded ones.  The
parameters of the model are learned from question-answer pairs.  For
example, the question \emph{What language do people in Czech Republic
  speak?} paired with its answer $\{\textsc{CzechLanguage}\}$. In line
with most work on question answering against Freebase, we do not rely
on annotated logical forms associated with the question for training
and treat the mapping of a question to its grounded graph as latent.

Let $q$ be a question, let $p$ be a paraphrase, let $u$ be an
ungrounded graph for $p$, and let $g$ be a grounded graph formed by
grounding the nodes and edges of $u$ to the knowledge base
$\mathcal{K}$ (throughout we use Freebase as the knowledge base).
Following \newcite{reddy_largescale_2014}, we use beam search to find
the highest scoring tuple of paraphrase, ungrounded and grounded
graphs $(\hat p, \hat u, \hat g)$ under the model $\theta \in \mathbb{R}^n$:
\[
({\hat{p},\hat{u},\hat{g}}) = \argmax_{(p,u,g)} \theta \cdot
\Phi(p,u,g,q,\mathcal{K})\,,
\]
where $\Phi(p, u, g, q, \mathcal{K}) \in \mathbb{R}^n$ denotes the
features for the tuple of paraphrase, ungrounded and grounded graphs.
The feature function has access to the paraphrase, ungrounded and
grounded graphs, the original question, as well as to the content of
the knowledge base and the denotation $|g|_\mathcal{K}$ (the
denotation of a grounded graph is defined as the set of entities or
attributes reachable at its \textsc{target} node).  See
\Cref{sec:details} for the features employed.  The model parameters
are estimated with the averaged structured perceptron
\cite{collins_discriminative_2002}.  Given a training question-answer
pair $(q,\mathcal{A})$, the update is:
\[
\theta^{t+1} \leftarrow \theta^{t} + \Phi(p^+, u^+, g^+, q,
\mathcal{K}) - \Phi(\hat{p}, \hat{u}, \hat{g}, q, \mathcal{K})\,,
\]
where $({p^+,u^+,g^+})$ denotes the tuple of gold paraphrase, gold
ungrounded and grounded graphs for $q$.  Since we do not have direct
access to the gold paraphrase and graphs, we instead rely on the set
of \emph{oracle tuples}, $\mathcal{O}_{\mathcal{K}, \mathcal{A}}(q)$,
as a proxy:
\[
(p^{+},u^{+},{g^{+}}) = \argmax_{(p,u,g) \in
  \mathcal{O}_{\mathcal{K},\mathcal{A}}(q)} \theta \cdot
\Phi({p,u,g,q,\mathcal{K}})\,,
\]
where $\mathcal{O}_{\mathcal{K}, \mathcal{A}}(q)$ is defined as the
set of tuples ($p$, $u$, $g$) derivable from the question $q$, whose
denotation $|g|_\mathcal{K}$ has minimal $F_1$-loss against the gold
answer $\mathcal{A}$.  We find the oracle graphs for each question a
priori by performing beam-search with a very large beam.

\section{Experimental Setup}
\label{sec:exp}

Below, we give details on the evaluation dataset and baselines used
for comparison. We also describe the model features and provide
implementation details.

\subsection{Evaluation Data and Metric}
\label{sec:evaldata}

We evaluate our approach on the \webq dataset \cite{berant_semantic_2013}.
\webq consists of 5,810 question-answer pairs where questions
represents real Google search queries. We use the standard train/test
splits, with 3,778~train and 2,032~test questions. For our development
experiments we tune the models on held-out data consisting of
30\%~training questions, while for final testing we use the complete
training data. We use average precision (avg P.), average recall (avg
R.) and average F$_1$ (avg F$_1$) proposed by
\newcite{berant_semantic_2013} as evaluation
metrics.\footnote{\url{https://github.com/percyliang/sempre/blob/master/scripts/evaluation.py}}


\subsection{Baselines}

\paragraph{\textsc{original}} 
We use \gp without paraphrases as our baseline. This gives an idea about the impact
of using paraphrases.

\paragraph{\textsc{mt}} 
We compare our paraphrasing models with monolingual machine
translation based model for paraphrase generation
\cite{quirk-2004,wubben:2010}. In particular, we use Moses
\cite{Koehn-2007} to train a monolingual phrase-based MT system on the
Paralex corpus. Finally, we use Moses decoder to generate
$10$-best distinct paraphrases for the test questions. 

\subsection{Implementation Details}
\label{sec:details}
 
\paragraph{Entity Resolution}

For \webq, we use 8~handcrafted part-of-speech patterns (e.g., the
pattern {\tt (DT)?(JJ.?$\mid$NN.?)\{0,2\}NN.?} matches the noun phrase
\textit{the big lebowski}) to identify candidate named entity mention
spans.  We use the Stanford CoreNLP caseless tagger for part-of-speech
tagging \cite{manning-2014-corenlp}.  For each candidate mention span,
we retrieve the top 10 entities according to the Freebase
API.\footnote{\scriptsize\url{http://developers.google.com/freebase/}}
We then create a lattice in which the nodes correspond to
mention-entity pairs, scored by their Freebase API scores, and the
edges encode the fact that no joint assignment of entities to mentions
can contain overlapping spans. We take the top 10 paths through the
lattice as possible entity disambiguations. For each possibility, we
generate $n$-best paraphrases that contains the entity mention
spans. In the end, this process creates a total of
$10n$~paraphrases. We generate ungrounded graphs for these paraphrases
and treat the final entity disambiguation and paraphrase selection as
part of the semantic parsing problem.\footnote{To generate ungrounded
  graphs for a paraphrase, we treat each entity mention as a single
  word.}

\paragraph{\gp Features.}  
We use the features from \newcite{reddy_largescale_2014}.  These
include edge alignments and stem overlaps between ungrounded and
grounded graphs, and contextual features such as word and grounded
relation pairs.  In addition to these features, we add two new real-valued
features -- the paraphrase classifier's score and the entity
disambiguation lattice score.

\paragraph{Beam Search} 
We use beam search to infer the highest scoring graph pair for a
question.  The search operates over entity-entity edges and entity
type nodes of each ungrounded graph.  For an entity-entity edge, there
are two operations: ground the edge to a Freebase relation, or skip
the edge. Similarly, for an entity type node, there are two
operations: ground the node to a Freebase type, or skip the node. We
use a beam size of 100 in all our experiments.






\section{Results and Discussion}
\label{sec:results}

In this section, we present results from five different systems for
our question-answering experiments: \textsc{original}, \textsc{mt},
\textsc{naive}, \textsc{ppdb} and \textsc{bilayered}. First two are
baseline systems. Other three systems use paraphrases
generated from an L-PCFG grammar. \textsc{naive} uses a word
lattice with a single start-to-end path representing the input
question itself, \textsc{ppdb} uses a word lattice constructed
using the PPDB rules, and \textsc{bilayered} uses bi-layered
L-PCFG to build word lattices. Note that \textsc{naive} does not
require any parallel resource to train, \textsc{ppdb} requires an
external paraphrase database, and \textsc{bilayered}, like
\textsc{mt}, needs a parallel corpus with paraphrase pairs. We tune
our classifier features and \gp features on the development data. We
use the best setting from tuning for evaluation on the test data.

\paragraph{Results on the Development Set}

Table~\ref{tab:devResults} shows the results with our best settings on
the development data. We found that oracle scores improve
significantly with paraphrases. \textsc{original} achieves an oracle
score of $65.1$ whereas with paraphrases we achieve an F$_1$ greater
than $70$ across all the models. This shows that with paraphrases we
eliminate substantial mismatch between Freebase and ungrounded
graphs. This trend continues for the final prediction with the
paraphrasing models performing better than the \textsc{original}.

All our proposed paraphrasing models beat the \textsc{mt}
baseline. Even the \textsc{naive} model which does not use any
parallel or external resource surpass the \textsc{mt} baseline in the
final prediction. Upon error analysis, we found that the \textsc{mt}
model produce too similar paraphrases, mostly with only inflectional
variations. For the question \textit{What language do people in Czech
  Republic speak}, the top ten paraphrases produced by \textsc{mt} are
mostly formed by replacing words \textit{language} with
\textit{languages}, do with \textit{does}, \textit{people} with
\textit{person} and \textit{speak} with \textit{speaks}. These
paraphrases do not address the structural mismatch problem. In
contrast, our grammar based models generate syntactically diverse
paraphrases.

Our \textsc{ppdb} model performs best across the paraphrase models
(avg F$_1$ = $47.9$). We attribute its success to the high quality
paraphrase rules from the external paraphrase database. For the
\textsc{bilayerd} model we found 1,000~latent semantic states is not
sufficient for modeling topical differences. Though \textsc{mt}
competes with \textsc{naive} and \textsc{bilayered}, the performance
of \textsc{naive} is highly encouraging since it does not require any
parallel corpus. Furthermore, we observe that the \textsc{mt} model
has larger search space. The number of oracle graphs -- the number of
ways in which one can produce the correct Freebase grounding from the
ungrounded graphs of the given question and its paraphrases -- is
higher for \textsc{mt}~(77.2) than the grammar-based models~(50--60).

\paragraph{Results on the Test Set}

Table~\ref{tab:mainResults} shows our final results on the test
data. We get similar results on the test data as we reported on the
development data. Again, the \textsc{ppdb} model performs best with an
F$_1$ score of $47.7$. The baselines, \textsc{original} and
\textsc{mt}, lag with scores of $45.0$ and $47.1$, respectively. We
also present the results of existing literature on this dataset. Among
these, \newcite{berant_semantic_2014} also uses paraphrasing but
unlike ours it is based on a template grammar (containing 8~grammar
rules) and requires logical forms beforehand to generate
paraphrases. Our \textsc{ppdb} outperforms Berant and Liang's model
by~7.8~F$_1$ points. 
\newcite{yih_semantic_2015} and \newcite{kun_question_2016} use neural network models for semantic parsing, in addition to using sophisticated entity resolution \cite{yang_smart_2015} and a very large unsupervised corpus as additional training data. Note that we use \gp as our semantic
parsing framework for evaluating our paraphrases extrinsically. We
leave plugging our paraphrases to other existing methods and other
tasks for future work. 

\paragraph{Error Analysis} The upper bound of our paraphrasing methods
is in the range of 71.2--71.8. We examine the reason where we lose the
rest. For the \textsc{ppdb} model, the majority~(78.4\%) of the errors
are partially correct answers occurring due to incomplete gold answer
annotations or partially correct groundings. Note that the partially correct groundings may include incorrect paraphrases. 13.5\%~are due to mismatch between Freebase and the paraphrases produced, and the rest~(8.1\%) are due to wrong entity annotations.

\begin{table}
  \small
  \begin{tabular}{lx{0.7in}x{0.5in}x{0.5in}}
    \toprule
    Method & avg oracle F$_1$ & \# oracle graphs & avg F$_1$ \\
    \midrule
    \textsc{original} & 65.1 & 11.0 & 44.7 \\
    \textsc{mt}  & 71.5 & 77.2 & 47.0 \\
    \textsc{naive} & 71.2 & 53.6 & 47.5 \\
    \textsc{ppdb} & 71.8 & 59.8 & 47.9 \\
    \textsc{bilayered} & 71.6 & 55.0 & 47.1 \\
    \bottomrule
  \end{tabular}
  \caption{\small Oracle statistics and results on the \webq
    development set.}
\label{tab:devResults}
\vspace{-0.4cm}
\end{table}

\begin{table}
  \centering
  \small
  \begin{tabular}{l|ccc}
    \hline 
    Method & avg P. & avg R. & avg F$_1$ \\ 
    \hline
    Berant and Liang '14\nocite{berant_semantic_2014} & 40.5 & 46.6 & 39.9 \\
    Bast and Haussmann '15\nocite{bast_more_2015}       & 49.8 & 60.4 & 49.4 \\
    Berant and Liang '15\nocite{berant_imitation_2015}& 50.4 & 55.7 & 49.7 \\
    Reddy et al. '16\nocite{reddy_transforming_2016} & 49.0 & 61.1 & 50.3 \\
    Yih et al. '15\nocite{yih_semantic_2015} & 52.8 & 60.7 & 52.5 \\
    Xu et al. '16\nocite{kun_question_2016} & 53.1 & 65.5 & 53.3 \\
    \hline
    \multicolumn{4}{c}{This paper} \\
    \hline
    \textsc{original} & 53.2 & 54.2 & 45.0 \\
    \textsc{mt} & 48.0 & 56.9 & 47.1 \\
    \textsc{naive} & 48.1 & 57.7 & 47.2 \\
    \textsc{ppdb} & 48.4 & 58.1 & 47.7 \\
    \textsc{bilayered} & 47.0 & 57.6 & 47.2 \\ 
    \hline
  \end{tabular}
  \caption{\small Results on \webq test dataset.}
  \label{tab:mainResults}
  \vspace{-0.4cm}
\end{table}

\section{Conclusion}
\label{sec:conclusion}

We described a grammar method to generate paraphrases for questions,
and applied it to a question answering system based on semantic
parsing. We showed that using paraphrases for a question answering
system is a useful way to improve its performance. Our method is
rather generic and can be applied to any question answering system.

\section*{Acknowledgements}

The authors would like to thank Nitin Madnani for his help with the
implementation of the paraphrase classifier. We would like to thank
our anonymous reviewers for their insightful comments. This research
was supported by an EPSRC grant (EP/L02411X/1), the H2020 project SUMMA
(under grant agreement 688139), and a Google PhD Fellowship for the second author.







\bibliographystyle{naaclhlt2016}

\bibliography{gen-qa}

\end{document}